\documentclass[twoside,11pt]{article}
\usepackage{jmlr2e}
\usepackage{times}
\usepackage{graphicx} 
\usepackage{subcaption} 

\usepackage{natbib}
\usepackage{color}
\usepackage{float}
\usepackage{algorithm}
\usepackage{algorithmic}
\usepackage[utf8]{inputenc} 
\usepackage{graphicx}
\usepackage{amssymb}
\usepackage{amsmath}
\usepackage{hyperref}
\usepackage[T1]{fontenc} 
\usepackage{tabularx}

\newtheorem{teo}{Theorem}[section]

\newtheorem{es}{Example}[section]

\newtheorem{defi}{Definition}[section]
\newtheorem{oss}{Remark}

	\def\R{\mathbb{R}}
	\def\P{\mathbb{P}}
	\def\E{\mathbb{E}}

\begin{document}

\title{Clustering Patients with Tensor Decomposition}

\author{\name Matteo Ruffini   \email mruffini@cs.upc.edu \\
       \addr Department of Computer Science\\
       Universitat Politècnica de Catalunya\\
       Barcelona, Spain.
       \AND
       \name Ricard Gavaldà   \email gavalda@cs.upc.edu \\
       \addr Department of Computer Science\\
       Universitat Politècnica de Catalunya\\
       Barcelona, Spain.
              \AND
       \name Esther Lim\'on  \email  elimon.mn.ics@gencat.cat \\
       \addr Institut Catal\`a de la Salut\\
       EAP Matar\'o-7, Barcelona, Spain}
\maketitle

\begin{abstract} 
In this paper we present a method for the unsupervised clustering of high-dimensional binary data, with a special focus on electronic healthcare records. We present a robust and efficient heuristic to face this problem using tensor decomposition. We present the reasons why this approach is preferable for tasks such as clustering patient records, to more commonly used distance-based methods. We run the algorithm on two datasets of healthcare records, obtaining clinically meaningful results.

\end{abstract} 

\section{Introduction}
\label{sec:intro}

Clustering patients in groups with similar clinical profiles is a strategic activity for modern healthcare systems. 
First, similar patients share the need for similar cares, so the system can design specific guidelines to treat and prescribe 
diagnostic patterns rather than single diagnostics. Also, clear and tested clusters based on comorbidities help clinicians 
to decide treatments on specific patients. Third, characterizing patient patterns helps the system in  planning resources and in performing fair comparisons  between institutions.  The context of this paper is a project that aims at making analytic techniques such as clustering accessible 
to healthcare managers, within a tool that allows them to explore and understand the population
under their care. We are working both with medical directors at large hospitals, with tens of thousands
of admissions per year, but also with planners and policy-makers at regional healthcare agencies providing health services
to a few million people. These professionals have little or no training in data science; their organizations often have a statistics department, but asking the department for analysis usually takes weeks, which hinders intuition and innovation, and leads in most cases to abandoning the analysis because of more pressing day-to-day matters. The goal of our project is to  make the algorithms in the tool largely autonomous, requiring no ad-hoc tuning before every analysis required by the user,
and efficient so that intuitions can be explored almost in real time, 
even for large populations. In the case of clustering, this setting clearly affects the algorithms that could be used. { A literature scan for clustering applications shows that most of the examples that can be found in the literature use standard \textit{distance based} clustering methods, like $k$-means \citep[see for example][]{rixen1996sepsis,kshetri2011modelling,perez2016rehabilitation},  or linkage-based methods  \cite[see ][]{cohen2010identification}. These techniques are general purpose algorithms based on the notion of distance, a concept that may lose part of its meaning in high dimensional settings \citep[see][]{aggarwal2001surprising,kriegel2009clustering},  especially when data are sparse and/or categorical as we anticipate to be the case. Additionally, the ``right'' notion of distance may be different for every application of the algorithm, i.e., may change if the manager is trying to cluster patients with diabetes, young or older population or any other possible profile. Furthermore, designing a distance function is too difficult or error-prone for our intended users, as it may involve non-trivial feature selection and engineering activity.
\\An alternative line of work uses nonnegative tensor factorization for phenotyping and clustering patients \citep[see][]{ho2014marble, ho2014limestone,wang2015rubik, 
perros2017spartan}; the idea is to use several patient-associated sources of data (called modes), like diagnostics, medical procedures, age, to create a tensor that, for each patient, counts the co-occurrences of the observations along the various modes. Decomposing this tensor using nonnegative low-rank decomposition methods one finds~$k$ vectors of phenotypes summing up the behavior of the modes in the dataset, and assigns each patient to the cluster/phenotype that most characterizes his/her records. This approach does not assume any statistical generative model for the data, requires to work with several sources of information and needs a high degree of manual interaction to define the modes and interpret the results.}\\
A different class of clustering techniques relies on the distributional properties of the observed variables, using  mixture models \citep[see][]{marin2005bayesian,melnykov2010finite}. Mixture models describe the observable data as outcomes of joint probability distributions and present many advantages over the cited techniques: they do not need an a-priori defined distance, { their nature of probabilistic graphical models allows natural and objective interpretations and their flexibility allows to work with both single and multiple sources of data}. In recent time, learning such models has
received attention using {\em methods of moments} together with {\em tensor decomposition} techniques \citep[see][and the references therein]{TensorLatent}. While traditional techniques, like pure Expectation Maximization (EM) \citep{EMDempster}, tend to scale too badly for the kind of interactive high-dimensional discovery that we envision, and often perform poorly, methods of moments provide guarantees of speed, stability and results quality that make them suitable for our purposes. This is the approach we take here, focusing on the case of mixture models where all the observable features are conditionally independent binary variables (mixture of independent Bernoulli).
We are not aware of any previous application of such methods to patient clustering.   In broad terms, these methods work by constructing from a sample some low-rank tensors representing the correlations
among the data attributes, then decomposing them to recover the parameters of the mixture model that explains these correlations
from a hidden variable. The key obstacle for the mixture of independent Bernoulli tends to be the first step, as there are several well-understood methods for the second. The problem of recovering the moments for mixtures with categorical features, was solved by \citet{jain2014learning} using an optimization algorithm, an approach that reduces the scalability of the method (the technique depends on a factor $d^3$ for both memory and time requirements, where~$d$ is the number of features). An alternative approach for these kind of data may be the so-called multi-view approach from \citet{TensorLatent}; however this method is prone to produce unreliable results when the data are unbalanced, with some few features that have more weight than the others, requiring in this case a nontrivial preprocessing of the data. We refer to the appendix for a detailed discussion on the related works on spectral methods.


\subsection{Our Contributions}

The contributions of this paper are the following: 

\begin{itemize}
\item We advocate for the use of methods moments for clustering datasets of patients. These methods can provide meaningful results, require little or no parameter tuning, do not need ad-hoc distance functions  and scale in existing
implementations to databases with at least tens of thousands of patients on commodity machines (few seconds on our example datasets, making interactive analysis a serious possibility). Also, unlike distance-based methods, they should not degrade when many irrelevant attributes are present.
\item We present an heuristic based on the method of moments to recover the parameters of a mixture of independent Bernoulli. The proposed method, instead of using a tensor-decomposition algorithm 
to retrieve asymptotically convergent estimates of the unknown parameters, uses this decomposition procedure to generate approximate parameters that are then used to feed EM as starting point. In the practical applications, using this technique, EM converges fast, reaching in few steps surprisingly good optima; also the high scalability allows the usage of this method on high-dimensional datasets.  While the main story of this paper would go through equally with several of the existing moments-based methods, the one presented here has a number of advantages (see appendix \ref{sec:other methods} for a deeper comparison): it requires the desired number of clusters as sole input, it scales as $d^2$ (where $d$ is the number of features) it only requires $d\geq k$ (where $k$ is the number of clusters - the multi-view approach requires for example $d \geq 3k$), and is deterministic, not randomized --- non-technical users distrust
algorithms that provide different results on different runs.


\item Finally, we apply our method to two real-world datasets obtained by collaboration with a healthcare agency.
The datasets contain data from hospital admissions of two selected profiles: patients with Congestive Heart Failure, and
``Tertiary'' patients (patients with serious diseases that can only be treated by top specialists or with highly specialized equipment).
The data contains essentially the list of ICD9 codes of the diagnostics at hospital admission time.
Data is turned into a binary matrix where columns represent diagnostics, so highly sparse, as every patient has most often less than a dozen diagnostics out of several thousand possible ones. 
The proposed method finds well-characterized clusters in the data, which clinicians
find meaningful, novel to some extent, and potentially useful for  refining clinical guidelines. 

\end{itemize}
Outline of the paper: Section \ref{sec:NBC} recalls how to use Na{\"\i}ve Bayes Models to perform clustering. 
Section \ref{sec:learn} presents the proposed learning  method and Section \ref{sec:patient clustering} contains our results on patient clustering.

\section{Na{\"\i}ve Bayes Clustering}\label{sec:NBC}
In this section we characterize the Na{\"\i}ve Bayes Model (NBM), together with an indication on how to use it to cluster data. Even if we are mainly interested on binary data, we keep the terms more general, as the presented results apply to all NBMs.
\begin{defi}\rm
A Na{\"\i}ve Bayes model (NBM) is a set of $d+1$ variables $(Y,x_1,\ldots,x_d)$ where $Y$ is a hidden (unobservable) discrete variable with a finite number of possible outcomes: $Y\in \{1,\ldots,k\}$. We define $
 \omega_h := \P(Y=h)$ and $\omega := (\omega_1,\ldots,\omega_k)^\top.$
The vector $X = (x_1,\ldots,x_d)$ is observable, its outcomes depend on the value of the hidden variable $Y$ and the random variables $x_1,\ldots,x_d$ are conditionally independent given $Y$; we define
 $\mu_{i,j} = \E(x_i|Y=j)$, and $M = ( \mu_{i,j})_{i,j}\in\R^{d\times k}.$
 Also we will denote with $\mu_i$ the set of columns of $M$: $M = [\mu_1|,...,|\mu_k].$
\end{defi}
The generative model first generates an unobservable  outcome for $Y$, say $h$, and then generates a set of observations $x_1,\ldots,x_d$, depending on $h$. If the parameters $(M,\omega)$ of a NBM are \textit{known}, they can be used for clustering purposes, assigning an  observation to the component with the highest probability of having generated it.
 
 \begin{es}[Mixture of independent Bernoulli]\label{es:mixtbin}\rm
 When the observables $x_i$ are binary, their conditional expectations coincide with the probability of a positive outcome:
 $
 \mu_{i,j}~=~\P(x_i~=~1|Y~=~j).
 $
In this fashion, we can derive a closed form of the probability that a given observation $X$ has been generated by a given hidden variable realization $Y=j$, $\P(Y = j |X)$, and perform clustering by assigning $Cluster(X) = argmax_{j = 1,...,k}(\P(Y = j |X))$, where, in this case:
 \begin{equation}\label{eq:mixtbern}
\P(Y = j |X) \propto \omega_j \prod_{i=1}^{d}\mu_{i,j}^{x_i}(1-\mu_{i,j})^{1-x_i}
\end{equation}
 \end{es}
Such Na{\"\i}ve Bayes Clustering procedure requires to know the conditional distribution of the observable data and the model parameters $({M},\omega)$. However, in the real world, these parameters are never known, and need to be estimated from a sample of observations of the observable features. In all the paper, we will assume to deal with a dataset $ \bar{X}\in \R^{N\times d}$ of size $N$ of observations denoted as:
$$
(\bar{X})_i = X^{(i)} = (x_1^{(i)},\ldots,x_d^{(i)}),\,\,\,i\in \{1,...,N\}.
$$ 
In the next sections we will present a method of moments to infer the pair  $({M},\omega)$ from a sample.

\section{Learning NBM parameters}\label{sec:learn}

When learning the parameters of a Latent Variable Model (LVM) with a tensor method, the typical approach consists in two steps. First we need to manipulate the observable moments in order to express them in the form of symmetric-low rank tensors, to approximate the following operators:
\begin{equation}
M_1 := \sum_{i=1}^{k}\omega_{i}\mu_{i},\,\,\,\, 
    M_2 := \sum_{i=1}^{k}\omega_{i}\mu_{i}\otimes\mu_{i},\,\,\,\,
M_3 := \sum_{i=1}^{k}\omega_{i}\mu_{i}\otimes\mu_{i}\otimes\mu_{i}\label{eq:M23}
\end{equation}
where $M_1 \in \R^{d}$, $M_2 \in \R^{d\times d}$ and $M_3 \in \R^{d\times d\times d}$ and $\otimes$ denotes the Kronecker product.
Second, we have to use a tensor decomposition algorithm to get $(M,\omega)$ from the equations \eqref{eq:M23}. 
{ The usage of the third order moments is necessary to guarantee the recoverability of $(M,\omega)$ with minimal assumptions on $M$ (like $M$ to be full rank). Methods dealing only with up to the second order moments exist, but present stronger requirements on the model parameters \citep{arora2013practical}. }
\subsection{Overview of the proposed method}\label{sec:overview}
The classic approach to learn the parameters of a LVM consists in designing asymptotically (with the sample size) convergent estimates of $M_1,M_2$ and $M_3$, and then to feed a tensor decomposition algorithm $\mathcal{A}$, to get asymptotically convergent estimates of  $(M,\omega)$. Unfortunately, the only method we are aware of to estimate $M_2$ and $M_3$ for a NBM with binary outcomes is that of \citet{jain2014learning}, that requires to deal with full tensors $M_3$, compromising the scalability of the method. In this paper instead we will follow a different approach: we will find two estimators whose expectations $\tilde{M}_2$ and $\tilde{M}_3$ are close, but not equal, to $M_2$ and $M_3$ (Subsection \ref{sec:mom}), and obtain an approximate estimate $(\tilde{M},\tilde{\omega})$ of $(M,\omega)$ using a tensor decomposition method (Subsection \ref{sec:svtd}). The usage of the proposed approximated estimators  $\tilde{M}_2$ and $\tilde{M}_3$ is crucial in order to allow the optimized implementation proposed in Appendix \ref{sec:embed}, that allows to recover the parameters $(\tilde{M},\tilde{\omega})$ without ever calculating explicitly $\tilde{M}_3$, which is not possible with the method from  \citet{jain2014learning}. As the estimators $\tilde{M}_2$ and $\tilde{M}_3$ are biased, also the retrieved parameters $(\tilde{M},\tilde{\omega})$ will suffer of a systematic bias; to get rid of it, we will refine the obtained parameters with EM,  (Subsection \ref{sec:EM}), increasing the likelihood of the model. The intuition behind this approach is that EM is able to refine the parameters that we get, removing the systematic bias, if this bias is small. This intuition is confirmed in the experiments of Appendix \ref{sec:expcomp}, where the proposed method outperforms many standard clustering methods when clustering binary data.

\subsection{Recovering the moments}\label{sec:mom}
We now focus on how to estimate $M_1$, $M_2$ and $M_3$ from a sample of observations distributed as a NBM. The estimation of $M_1$ is a trivial consequence of the conditional expectation: given a sample 
$
(\bar{X})_i = X^{(i)} = (x_1^{(i)},\ldots,x_d^{(i)}),$ for $i\in \{1,...,N\}
$ we have, for each $j$, $
(M_1)_j = \lim_{N\to \infty}{\sum_{i=1}^{N}{x_j^{(i)}}}/{N}.$
We now focus on $M_2$ and $M_3$. Define the following  quantities:
\begin{equation}
\tilde{M}_2^{(N)} = \frac{\bar{X}^\top \bar{X}}{N}\in \R^{d \times d},\,\,\,\,\,\,
\tilde{M}_3^{(N)} =  \sum_{i=1}^{N} \frac{X^{(i)} \otimes X^{(i)} \otimes X^{(i)}}{N}  \in \R^{d \times d \times d}\label{eq:M23t}
\end{equation}
and their respective limits,
$\tilde{M}_2 = \lim_{N\to\infty}\tilde{M}_2^{(N)}$ and $\tilde{M}_3= \lim_{N\to\infty}\tilde{M}_3^{(N)}.
$
It is possible to demonstrate (see Appendix \ref{sec:teoerr} for details) that $\tilde{M}_2$ and $\tilde{M}_3$ are close to $M_2$ and $M_3$; in particular, the estimates $\tilde{M}_2$ and $\tilde{M}_3$ always converge to the right value in the off-diagonal terms, while the diagonal entries instead converge to a value whose distance from the true value is bounded and can always be estimated from the sample. Even if biased, we will use $\tilde{M}_2$ and $\tilde{M}_3$ as input to a tensor decomposition algorithm, to estimate the parameters $(\tilde{M},\tilde{\omega})$, which will consequently be biased. To reduce this bias, once learned the parameters of a NBM, we will update them with some iterations of EM, in order to increase the likelihood of the learned model.

\subsection{Tensor Decomposition with SVTD}\label{sec:svtd}
Given the true values of \eqref{eq:M23}, we can use a decomposition algorithm to learn the NBM parameters. The literature provides a variety of algorithms to retrieve the desired pair $(M,\omega)$ from the estimated tensors, which differ one from another on how they exploit the spectral properties of the moment operators \citep{SpectralLDA,SpectralLatent, TensorLatent,kuleshov2015tensor}. Even if the choice of the decomposition algorithm to be used is arbitrary, in this paper, we will base on SVTD, from \citet{ruffini2016new}, for two reasons: first it is deterministic, not relying on any randomized matrix for its implementation; second, because it allows, with few simple modifications, to work without dealing explicitly with the tensor $M_3$, that is often too large (cubic in the number of variables); these modifications, which reduce the complexity of the method, are presented in Appendix \ref{sec:embed} where we embed the estimation procedure of the previous Subsection \ref{sec:mom} with SVTD. A summary of the basis of SVTD, as presented by \citet{ruffini2016new} can be found in Appendix \ref{sec:svtdappendix}. When fed with the correct model parameters $M_1$, $M_2$, $M_3$, SVTD produces the correct estimates $(M,\omega)$. Conversely, when the inputs are biased, also the estimated $(\tilde{M},\tilde{\omega})$ will suffer of a bias that depends on the bias of the inputs \citep[][]{ruffini2016new}.
To remove this bias, we will run few iterations of EM.
\subsection{Expectation Maximization}\label{sec:EM}
In previous sections we have seen how to compute from data an estimate $(\tilde{M},\tilde{\omega})$ of the parameters of a NBM. We can improve that estimate using \textbf{E}xpectation \textbf{M}aximization (EM)  \citep{EMDempster}. EM, starting from an initial estimate  $(\tilde{M},\tilde{\omega})$, iteratively updates those parameters, increasing at each step the likelihood of the model, until convergence is reached. The issue of EM is that it may converge to poor local optima, if initialized poorly. Even if the results provided by the method described here suffer for the bias introduced in the estimates of moments, for sufficiently large datasets, those estimates are close to the correct estimates and so also the outputs will be close to the "correct" results \citep[see][]{ruffini2016new}; the intuition is that the biased pair  $(\tilde{M},\tilde{\omega})$ of the previous paragraph, are  good initializers for EM, and a few iterative steps will suffice to correct the initial bias, reaching a good optimum \citep[See][for details on EM]{marin2005bayesian}.
  
\section{Putting it all together} \label{sec:all}
We are now ready to merge all the elements presented till now, and to show how to use them to perform Na{\"\i}ve Bayes Clustering. Starting from a dataset $\bar{X}\in\R^{N \times d}$, one can follow the following flow: \textit{First}, estimate $\tilde{M_1}$, $\tilde{M_2}$, $\tilde{M_3}$, as in Subsection \ref{sec:mom} and retrieve the model parameters $(\tilde{M},\tilde{\omega})$ with SVTD. We call  \textit{Approximate-SVTD}, or \textit{ASVTD}, this  method built on approximated tensor decomposition, and we refer to Algorithm \ref{alg:SVTD approx} in Appendix \ref{sec:embed} for an optimized implementation { that, embedding the moments computation   in the decomposition algorithm, allows to run the tensor decomposition without explicitly computing $\tilde{M_3}$ allowing the method to scale as $d^2$ instead of $d^3$}. \textit{Then},  apply EM to improve the quality of the estimated parameters. \textit{Last}, use Na{\"\i}ve Bayes Clustering to cluster the rows of $\bar{X}$. 
These steps will be applied in the next section on two medical datasets.
{ \begin{oss}[Tuning the clustering results]\rm
The EM runs performed after the parameter retrieval, guarantee that the retrieved mixture model is (locally) optimal in terms of likelihood. Additional fine-tuning is possible re-running the algorithms in different configurations, for example modifying the number of clusters, adding/excluding certain features, 
or performing additional clustering within one of the clusters. The interpretation technique provided in the next section allows a user to visually inspect results and add expert considerations on the discovered patterns. Other type of manual fine-tuning (like manually assign a patient to a different cluster) may be possible, at the cost of adding human bias to the results of the model.
\end{oss}}

\section{Patient Clustering}\label{sec:patient clustering}

In this section we present the results of applying ASVTD to cluster patients form a real-world EHR\footnote{Due to privacy constraints, these datasets are not publicly available.}.

\subsection{The datasets}
We consider two datasets provided by
Servei Catal\`a de la Salut, the major provider of public healthcare in Catalonia, Spain. The datasets are subsets of a bigger database containing all hospitalizations in Catalonia for the year 2016. The first dataset contains all the patients affected by Heart Failure, to be precise patients having a primary or secondary diagnostic 428 in the ICD-9 code. The second dataset contains ``tertiary'' patients, those with a serious disease that is supposed to be treated in one of the 5-6 large, reference hospitals 
in the area, because either the expertise or the required 
resources are only found there; criteria for tertiarism includes all oncology surgery, all heart surgery, all neurosurgery,
major traumatism, acute ictus, myocardial infarct and sepsis, transplants, and a few rarer conditions.
The datasets have the same format: each row represents a visit of a patient to an hospital, and the columns contain the codes of the up to 10 diagnostics that the doctor annotated in the patient history at the time of admission; their order is considered irrelevant. Diagnostics are coded in the international ICD-9 code \citep{geraci1997international}. The codes are hierarchical, of the form ddd.dd, with the .dd part being optional:  the first 3 digits are a general diagnostic and extra digits if present make the diagnostic more precise. We kept only the first three digits of the ICD-9 codes, which are those with more consensus among doctors, obtaining a total vocabulary of 696 codes. 
In this way it is straightforward to map a dataset into a $N\times 696$ matrix with binary entries, where $N$ is the number of records of each dataset, using an approach analogous to the bag-of-words: first associate each registered disease to a unique number between 1 and 696, and then populate a matrix $\bar{X}$ placing a 1 at position $(i,j)$ if the record $i$ presents diseases $j$, otherwise a zero.

\subsection{Analysis of the results}
In both the cases we run ASVTD, followed by EM, to then perform Na{\"\i}ve Bayes Clustering; the stopping criteria for EM was the moment in which convergence was reached; in particular, we stopped the algorithm when the norm of the variation on the estimated $\omega$ between an iteration and the next was less than $0.01$. In both cases, to perform tensor decomposition, we removed the records with less than 3 diseases, obtaining two reduced datasets: for the Tertiarism dataset,  $\bar{X}_{t}$ with $N = 16311$ samples and  for the Heart Failure dataset $\bar{X}_{c}$ with $N= 23154$ samples. In order to analyze the results we plot two charts for each dataset: a heat-map and a disease-frequency chart.  The heat-map, is in figure \ref{fig:Cheatmap}. In that figure, we took the 40 most common diseases and plot the matrix $\bar{X}_{c}$ only for the columns associated to those diseases. We ordered the rows of $\bar{X}_{c}$ according to the cluster to which they belong, and draw them sequentially: each row of Figure \ref{fig:Cheatmap} is a record, the black points indicate whether a disease is present. The background color of each row indicates the cluster (so: white background is the first cluster, clear gray the second, and so on). The purpose of this figure is to give a first visual inspection of the patterns present inside the clusters; heat-maps also give an idea of the dimension of the clusters with respect to the total dataset. Disease-frequency charts are instead used to give a meaning to those patterns, an example is at Figure \ref{fig:CFreq}: the top-20 diseases in terms of frequency are displayed, each one in a different row. Then, for each one of the clusters, we show the relative frequencies of the considered diseases; for example, in  Figure \ref{fig:CFreq}, ``Dyslipidemia'' is present in the 50\% of the patients belonging to the first cluster. 
To have an additional indication of the content of the clusters, we represent in Tables \ref{tab:Heart} and \ref{tab:Tertiarism} the most  \textit{relevant} diagnostics, where the \textit{relevance}  \citep{sievert2014ldavis} of a diagnostic $i$ with respect to a cluster $j$, given a weight parameter $\lambda$, is defined as
$$
r(i,j) =  \lambda\log( \mu_{i,j})+(1-\lambda  )\log(  \mu_{i,j}/ (\sum_{h=1}^{k}\mu_{i,h}\omega_h))
$$
where $\mu_{i,j} = \P(x_i=1 | Y=j)$. The relevance of a diagnostic with respect to a cluster $j$ is an indicator that has a high value if the frequency of the considered diagnostic inside the cluster is much higher then its frequency on the full dataset; diagnostics with a high relevance are those that characterize a given cluster with respect to the others.
An analysis of the content of a cluster can be performed merging the information contained in the disease-frequency charts with those provided by the tables of the highly relevant diseases: disease-frequency charts will show us the diseases that are frequent in a given cluster, regardless what happens in the other clusters; conversely, tables of the highly relevant diseases will highlight those diagnostics that characterize a cluster with respect to the others, providing in this way a complete view on the patterns that can be found in the data.

\subsubsection{Heart failure dataset}
 \begin{figure*}[h!]\label{fig:Card}
    \centering
    \begin{subfigure}[b]{0.65\textwidth}
            \centering
            \includegraphics[width=\textwidth]{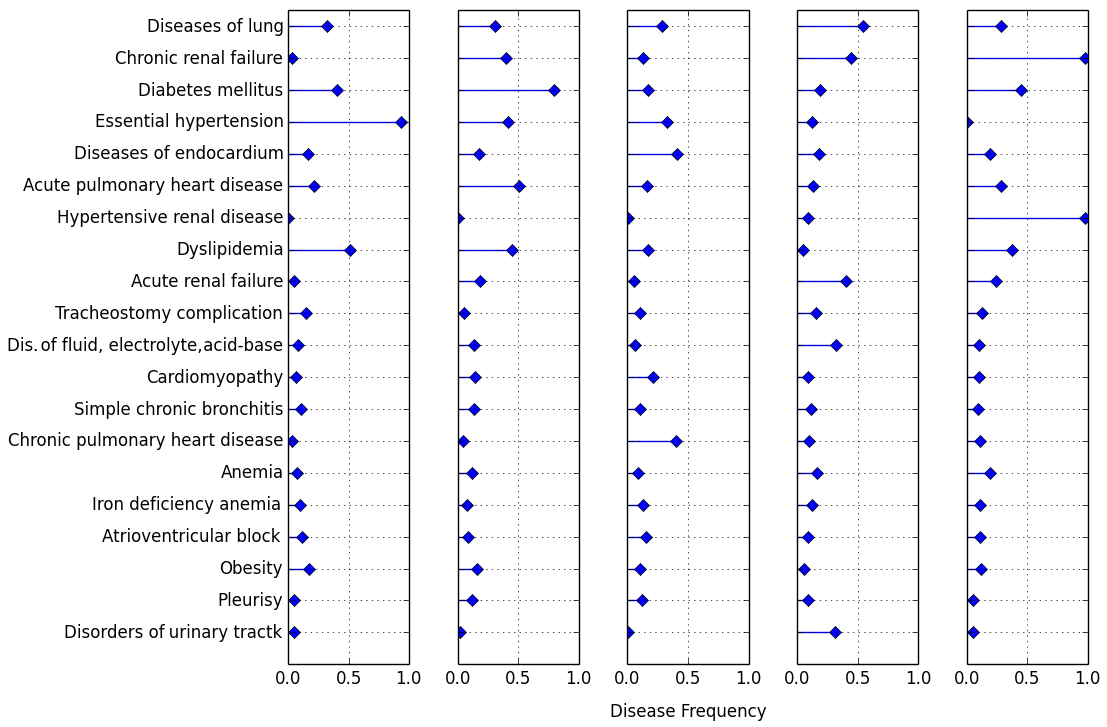}
            \caption{}
            \label{fig:CFreq}
    \end{subfigure}
\begin{subfigure}[b]{0.25\textwidth}
            \centering
            \includegraphics[height=7cm]{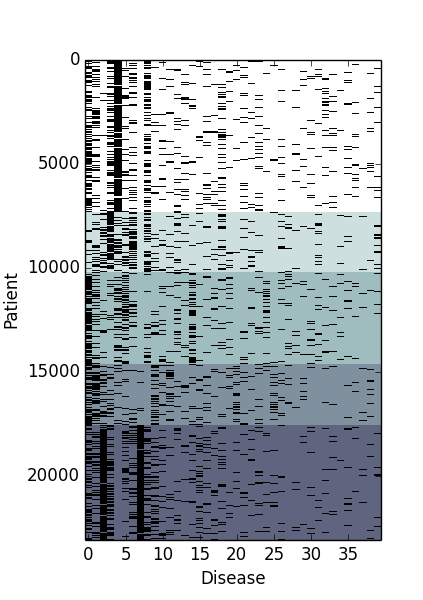}
            \caption{}
            \label{fig:Cheatmap}
    \end{subfigure}
\caption{\ref{fig:CFreq} Disease-frequency chart for the Heart Failure dataset. \ref{fig:Cheatmap} The corresponding heatmap. The ``Heart Failure'' disease is not represented as it appears in all the records of the dataset.}
\end{figure*}
\begin{table}[!t]\footnotesize
\centering
\begin{tabularx}{\textwidth}{c X c}
\hline
\textbf{Cluster} & \textbf{Most Relevant Diagnostics} & \textbf{Cluster Size}\\
\hline
1 & Essential hypertension;
Dyslipidemia;
Cardiac arrhythmias;
Diabetes mellitus;
Obesity;
  &7290
\\\hline
2 & Diabetes mellitus;
Atherosclerosis;
Acute pulmonary heart disease;
Retinal disorders;
Hypertensive heart and renal disease;
&  2915 
\\\hline
3 & Chronic pulmonary heart disease;
Diseases of endocardial structures;
Diseases of endocardium;
Hypertensive heart disease;
Cardiac arrhythmias;
 &  4480 
\\\hline
4 & Bacterial infection;
Disorders of urethra - urinary tractk;
Acute renal failure;
Hypertensive heart - renal disease;
Disorders of fluid, electrolyte, acid-base balance;
Diseases of lung;
& 2936
\\\hline
5 & Hypertensive renal disease;
Chronic and/or acute renal failure;
Diabetes mellitus;
Anemia;
& 5533
\\\hline
\end{tabularx}
\caption{The most relevant diseases for each cluster for the Heart failure dataset.}\label{tab:Heart}
\end{table}
The clusters obtained in the heart failure dataset are plotted in Figure \ref{fig:CFreq}, with the associated table \eqref{tab:Heart} of the highly relevant diseases. Clusters can be described with high coherence: Cluster~1 could be called the ``purely metabolic'': high prevalence of hypertension, and also dyslipidemia and diabetes mellitus.
Cluster 2 contains the above complicated with kidney problems;
Patients in Cluster 3 present valvular problems and pulmonary hypertension, well-known to be related.
Cluster~4 is a mixture of kidney and pulmonary problems (lung diseases are pretty frequent here), with little metabolic implications. Cluster 5 contains the purely kidney sufferers, with diabetes mellitus (nephropaty being a common complication of diabetes) and dyslipidemia. Clinicians confirm that these clusters make sense once seen, and may be useful for guiding treatment. For example, medication that might be indicated for the purely cardiac clusters might be not advisable for the clusters with renal problems if it is suspected to be nephrotoxic.

\subsubsection{Tertiarism dataset}
 \begin{figure*}[h!]\label{fig:Terc}
    \centering
    \begin{subfigure}[b]{0.65\textwidth}
            \centering
            \includegraphics[width=\textwidth]{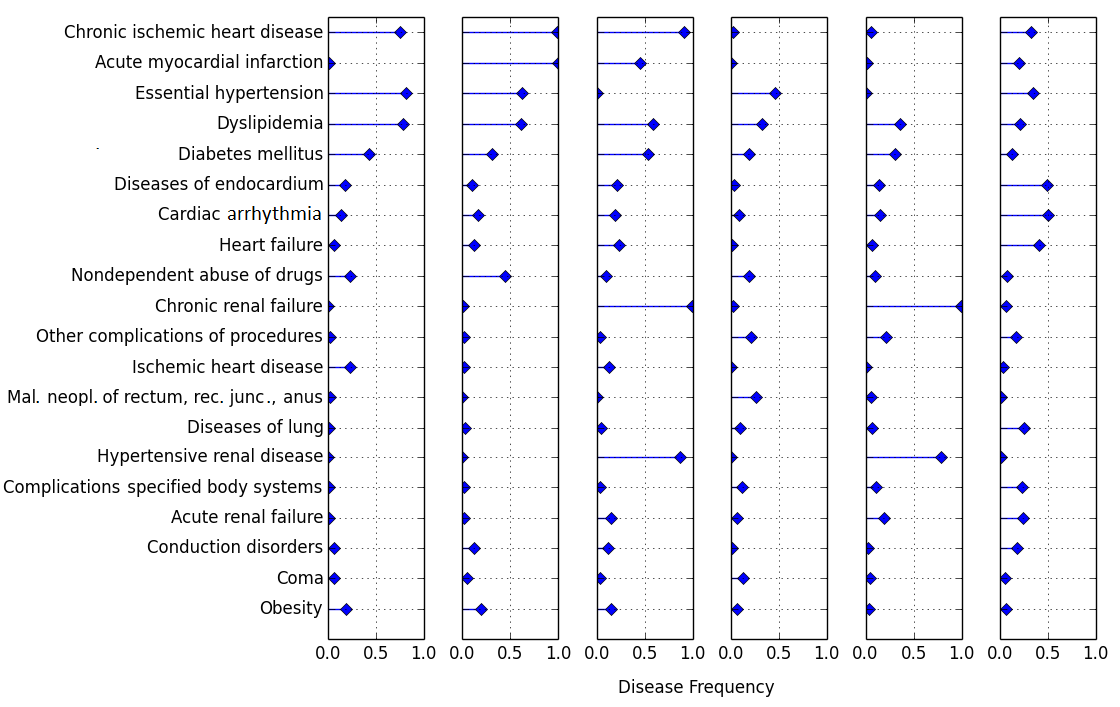}
            \caption{}
            \label{fig:TFreq}
    \end{subfigure}
\begin{subfigure}[b]{0.3\textwidth}
            \centering
            \includegraphics[height=6.5cm]{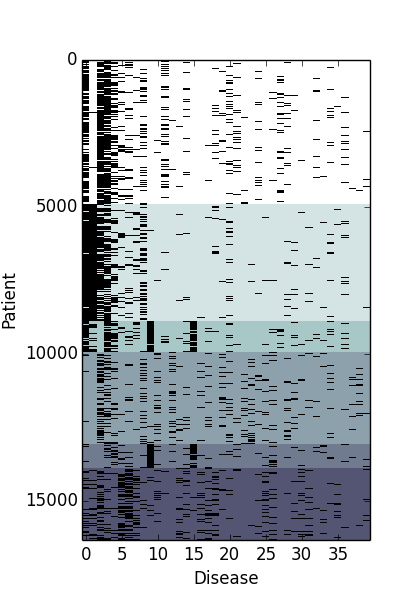}
            \caption{}
            \label{fig:Theatmap}
    \end{subfigure}
\caption{\ref{fig:TFreq} Disease-frequency chart for the Tertiarism dataset. \ref{fig:Theatmap} The corresponding heatmap. }
\end{figure*}

\begin{table}[!t]\footnotesize
\centering
\begin{tabularx}{\textwidth}{c X c}
\hline
\textbf{Cluster} & \textbf{Most Relevant Diagnostics} & \textbf{Cluster Size}\\
\hline
1 & Ischemic heart disease;
Essential hypertension;
Angina pectoris;
Dyslipidemia;
Chronic ischemic heart disease;
  &4892

\\\hline
2 & Acute myocardial infarction;
Chronic ischemic heart disease;
Nondependent abuse of drugs;
Essential hypertension;
Dyslipidemia;
&  3982 
\\\hline
3 & Hypertensive renal disease;
Chronic renal failure;
Chronic ischemic heart disease;
Diabetes mellitus;
Acute myocardial infarction;
 &  1043 
\\\hline
4 & Malignant neopl. of rectum, rectosigmoid junction, and anus;
Secondary malignant neopl. of respiratory and digestive systems;
Malignant neopl. of trachea, bronchus, and lung;
Malignant neopl. of stomach;
Diseases of white blood cells;
& 3133 
\\\hline
5 & Chronic renal failure;
Hypertensive renal disease;
Disorders involving the immune mechanism;
Anemia;
Disorders of urethra and urinary tractk;
& 819
\\\hline
6 & Diseases of endocardial structures;
Chronic pulmonary heart disease;
Heart failure;
Diseases of endocardium;
Diseases of lung;&  2442 
\\\hline
\end{tabularx}
\caption{The most representative diseases for each cluster for the Tertiarism dataset.}
\label{tab:Tertiarism}
\end{table}
The clusters for the tertiarism dataset  are plotted in Figure \ref{fig:TFreq}, with the associated Table \ref{tab:Tertiarism} of the high relevant diseases.
Clinically, Cluster 1 might correspond to patients in need of coronary intervention (heart surgery or interventional cardiology), again with strong presence of metabolic anomalies. Cluster 2 is similar, but including myocardial infarction. Cluster 3 includes nephrology patients, who are also taken to tertiary hospitals when they
need transplant or are in very advanced phase, and who 
tend to develop arteriopathy in the long term; interestingly,
diabetes mellitus shows up strongly in this cluster, as it is the leading cause of entrance to dialysis programs.
Cluster 4 has a surprising behavior: on a first hand, no clear signal comes out from the disease-frequency chart, besides the ever-present hypertension, dyslipidemia and a small signal of colo-rectum cancer. However, if we look at the table of high relevant diseases, we can see that the most relevant diagnostics for this cluster are neoplasms. Oncological problems are large constituents of tertiarism; they are characterized by a large set of codes (almost 100 for neoplasms, by organ of origin mostly), so it is not surprising the fact that we do not find such a diagnostic in the disease-frequency plot, as there are many distinct diagnostics indicating oncological diseases; however, the fact that the most relevant diagnostics for this cluster are neoplasms allows us to consider this family of diagnostics as characterizing this cluster. Cluster 5 groups kidney patients without arteriopathy, so similar to the first and second clusters but without previous infarction. Cluster 6 includes patients with cardiac valvular disease; normally they need heart surgery or interventional cardiology, but their associated complications are quite different from those in Clusters 1 and 2. 
\paragraph{Stability analysis.}A clustering method is expected to be stable: if trained on two datasets of reasonable size generated by the same process, it should produce similar clusters. To study the stability of ASVTD we split each of the studied datasets into two non-disjoint subsets on which we run the clustering method; then we compare the clusterings obtained on the intersection between the two subsets with the Adjusted Rand Index \citep{hubert1985comparing}. In both the cases, we got satisfying results, with an index higher than 0.9, indicating similarity between the two partitions and thus stability. 

\section{Conclusions}
We have presented an efficient method for clustering high dimensional binary data, able to find interesting and potentially useful patterns even in the simplest format of EHR, namely, diagnostic codes only. A strong point noted by clinicians is that it provides a fresh look, without prejudice, based on an aseptic algorithm.    A natural follow-up consists in including  additional clinical information to the features, like demographics, lab results, vital signs or medications. Also, we believe that including higher-level diagnostic groups among the features may improve the clarity of some patterns; these are more clearly defined in the more recent coding scheme ICD10 than in ICD9.

\acks{Work by M.\ Ruffini and R.\ Gavald\`a is partially funded by 
AGAUR project 2014 SGR-890 (MACDA) and by MINECO project TIN2014-57226-P (APCOM).
}

\newpage  

\bibliography{AR_biblio}  

\newpage  

\appendix 
\section*{\Large Appendix}

\section{Related works on spectral methods}
\label{sec:relatedwork}

A  few examples of application of mixture models to the medical domain can be found in the literature \citep[see][]{sun2007multivariate}, but we are not aware of any of them using methods of moments and tensor decomposition. We are interested in performing clustering using a special case of mixture model, called Na{\"\i}ve Bayes Model (NBM), where all the observable features are conditionally independent binary variables. 
The key obstacle to learn such models with a spectral method consists in recovering from a sample a good approximation of the low rank tensors to be decomposed.
In \citep{jain2014learning} the authors solve this problem for mixtures with categorical outcomes employing an optimization algorithm to reconstruct such tensors; then, they suggest to feed Tensor Power Method from  \citet{TensorLatent} to recover the model parameters. Their approach works when the number of the features is significantly higher than the number of the clusters; the problem  of that method is the fact that it requires the storage of the full three dimensional tensor in memory, reducing the scalability of the algorithm (depending on a factor $d^3$ for both memory and time requirements, where $d$ is the number of features). An alternative approach for some mixture models, including NBMs was proposed in \citep{SpectralLatent}; this approach, which requires the number of features to be at least three times the number of clusters ($d\geq 3k$), consists in treating a NBM as a three-views mixture model, by splitting the vector of the features into three subvectors, and independently learn the parameters of each one of the three views. The method proposed in that paper  relies on the usage of randomized matrices, that highly affect the quality of the retrieved parameters; some of the same authors, presented in \citep{TensorLatent} a different way to deal with the multi-view approach (and so with NBMs), together with the Tensor Power Method. This approach is robust, and seems to work fine in many different settings. We remark that it is possible to implement these techniques in an optimized way, reducing the complexity of the method to $d^2$, allowing the management of large scale datasets. 

\subsection{Why a different technique?}
\label{sec:other methods}

The main drawback of the multi-view approach is the fact that it requires the three views to be fully informative about the latent variable: geometrically, this means that they are required to be full-rank; practically, it means that we can infer the status of a latent variable just by looking at one single view. With healthcare electronic records, this may not be the case.
Consider a dataset of $N$ patients, with $d$ binary records. We can think at this dataset as a matrix $\bar{X}$ with $N$ rows and $d$ columns, where at position $(i,j)$ we have a 1 if the patient $i$ has the disease $j$. In general, the frequency of the various diseases along the population of patients has a huge tail, with few most common diseases that are extremely spread along the population, and a very high portion of the remaining diseases that are pretty rare; however, even if a disease is rare for the whole population it might be characteristic of a small cluster of patients. 
In such a context, to implement a multi-view approach, we need to split the columns of the matrix $\bar{X}$ into three groups, obtaining three datasets; each one would represent a view on the latent status of the patient, which could be inferred by looking at an individual view. In our preliminary experiments, we tested the multi-view method on our databases of medical records, and we obtained results that were highly dependent on the way in which the split of the columns was made. Finding the "right" view, resulted to be a tricky task, highly depending on the analyzed dataset. 
A second issue with the multi-view approach, regards the ordering issues that emerge from the asynchronous learning of the three views. Each view is a matrix whose column $j$ represents the conditional expectation of the features of the view under cluster $j$. In general, when the views are returned by a tensor decomposition algorithm, the ordering of the columns is not consistent between the various views, and to obtain a proper reordering  matching heuristics are required  \citep[see Section B.4 of][]{SpectralLatent}. With a broader view, for the specific application that we have in mind, one requires algorithms that: 1) need little hand-crafting (such as parameter tuning or data preprocessing like splitting the features into views, requiring long hours of collaboration between medical expert and data scientist), 2) are deterministic (doctors become highly suspicious of an algorithm that returns different results on different runs and same data), 3) have interpretable output and 4) are scalable, in the sense that can be run in seconds on large scale datasets. Besides the technical aspects on tensor decomposition method sketched above, our contribution is toward a clustering algorithm that is truly usable in this sense.

\section{On the estimated tensors or moments }\label{sec:teoerr}
In the following theorem we show that the estimates proposed in Section \ref{sec:mom} are close to the theoretical, unbiased ones.
 \begin{teo}\label{teo:eur}\rm
Define the values of the conditional second and third order moments for an observable variable:
 $$
 \mu^{(2)}_{i,j} = \E(x_i^2|Y=j),\,\,\,\,\, \mu^{(3)}_{i,j} = \E(x_i^3|Y=j).
 $$
 Then $(M_2)_{i,j} = (\tilde{M}_2)_{i,j}$ and $(M_3)_{i,j,l} = (\tilde{M}_3)_{i,j,l}$ for all $ i \neq j \neq l$. Also, if  $\mu_{i,j} \in [0,1]\,\, \forall i,j$, we have for all $ i \neq j \neq l$:
 \begin{align*}
 &|(M_2)_{i,i} - (\tilde{M}_2)_{i,i}|\leq\sum_{j=1}^{k}\omega_j\mu^{(2)}_{i,j} -  (M_1)_i^2,\\
|(M_3)_{i,i,l} - (\tilde{M}_3)_{i,i,l}|\leq &\sum_{j=1}^{k}\omega_j\mu^{(2)}_{i,j}\mu_{l,j} -\frac{(M_2)_{i,l}^2}{(M_1)_l}, \,\,\,
|(M_3)_{i,i,i} - (\tilde{M}_3)_{i,i,i}|\leq \sum_{j=1}^{k}\omega_j\mu^{(3)}_{i,j} -(M_1)_{i}^3.
  \end{align*}
 \end{teo}
Notice that, because $M_3$, $\tilde{M}_3$, $M_2$ and $\tilde{M}_2$ are symmetric, the previous theorem states an error bound on all the entries of that operators. 
\proof
Recall the following notation
$$
 \mu^{(2)}_{i,j} = \E(x_i^2|Y=j),\,\,\,\,\, \mu^{(3)}_{i,j} = \E(x_i^3|Y=j)
 $$
 and observe that, because each $ \mu_{i,j} \in [0,1]$ ,
 $$
  \mu^{(2)}_{i,j}\geq \mu^{2}_{i,j},\,\,\,\,\, \mu^{(3)}_{i,j}\geq \mu^{3}_{i,j}.
 $$
We prove each claimed equation one by one.

\begin{enumerate}
\item  $
 (M_2)_{i,j} = (\tilde{M}_2)_{i,j},$ and $(M_3)_{i,j,l} = (\tilde{M}_3)_{i,j,l} 
 $
  for all $ i \neq j \neq l$.
 These equations are an immediate consequence of the conditional independence of the features.
 
 \item  $
 |(M_2)_{i,i} - (\tilde{M}_2)_{i,i}|\leq\sum_{j=1}^{k}\omega_j\mu^{(2)}_{i,j} -  (M_1)_i^2
 $
 
 We consider 
 $$
 (\tilde{M}_2)_{i,i} = E(x_i^2) = \sum_{j=i}^k\omega_j\mu^{(2)}_{i,j}\geq \sum_{j=i}^k\omega_j\mu_{i,j}^2 = 
 $$
  $$
 =  (M_2)_{i,i} \geq (\sum_{j=i}^k\omega_j\mu_{i,j})^2 =(M_1)_i^2.
 $$
 From which we get the desired inequality.
 \item $
|(M_3)_{i,i,l} - (\tilde{M}_3)_{i,i,l}|\leq \sum_{j=1}^{k}\omega_j\mu^{(2)}_{i,j}\mu_{l,j} -\frac{(M_2)_{i,l}^2}{(M_1)_l} 
 $.
 We have
 $$
 (\tilde{M}_3)_{i,i,l} = \sum_{j=1}^{k}\omega_j\mu^{(2)}_{i,j}\mu_{l,j} \geq  \sum_{j=1}^{k}\omega_j\mu^{2}_{i,j}\mu_{l,j} = 
 $$
$$
= (M_3)_{i,i,l} \geq \frac{(M_2)_{i,l}^2}{(M_1)_l},
 $$
from which again we get the desired inequality.

 \item $
|(M_3)_{i,i,i} - (\tilde{M}_3)_{i,i,i}|\leq \sum_{j=1}^{k}\omega_i\mu^{(3)}_{i,j} -(M_1)_{i}^3 
 $.
 
 The proof of this point is identical to that of point 2.
 \endproof

\end{enumerate}

\section{Tensor Decomposition with SVTD: details}\label{sec:svtdappendix}
In this section we describe in depth the tensor decomposition method we used in the paper, SVTD. SVTD requires the matrix $M$ to have rank $k$, with at least one row where all the entries are distinct. It is based on two key observations: 1) the matrix $M_2$ has rank $k\leq d$, $k$ is the number of states that the latent variable can assume, and $M_2$ admits the following representation $M \Omega M^\top $, where $\Omega = diag(\omega)$; 2) for each $r \in \{1,...,d\}$, the $r-th$ slice of the tensor $M_3$ can be expressed as a rank$-k$ matrix with the following structure:
$$
M_{3,r} = M \Omega^{\frac{1}{2}}  diag((\mu_{r,1},...,\mu_{r,k}))\Omega^{\frac{1}{2}} M^\top
$$
  
The algorithm first performs a singular value decomposition (SVD) on $M_2$, $M_2 = U_k S_k U_k^\top,$ where $U_k\in\R^{d\times k}$ and $S_k  \in \R^{k\times k}$ are the matrices of the singular vectors and values truncated at the $k-th$ greatest singular value. Then defines
$
E_k = U_kS_k^{\frac{1}{2}} \in \R^{d\times k},
$
and for each slice $M_{3,r}$ of $M_3$, defines $H_r \in \R^{k\times k}$ as
$$
H_r = E_k^{\dagger} M_{3,r} (E_k^\top)^{\dagger},
$$
where $E_k^{\dagger} = (S_k)^{-\frac{1}{2}}U_k^\top $ is the Moore-Penrose pseudo-inverse of $E_k$. Observing that there exists an orthonormal $k\times k$ matrix $O$, such that 
$M \Omega^{\frac{1}{2}} =  E_kO $, one gets the following characterization of $M_{3,r} $:
$$
M_{3,r} = M \Omega^{\frac{1}{2}}  diag((\mu_{r,1},...,\mu_{r,k}))\Omega^{\frac{1}{2}} M^\top$$
$$= E_kO diag((\mu_{r,1},...,\mu_{r,k}))O^\top E_k^\top,
$$
from which it follows that
$$
H_r = O diag((\mu_{r,1},...,\mu_{r,k}))O^\top.
$$
Now, one gets the $r-$th row of $M$ as the singular values of $H_r$.
Repeating these steps for all the $r\in \{1,...,d\}$ will provide  the full matrix $M$. In order to avoid ordering issues with the columns of the retrieved $M$, one can 
use the same matrix $O$ to diagonalize all the matrices $H_r$, as $O$ does not depend on the considered row $r$. So, compute $O$ as the singular vectors of $H_r$, for a certain feature $r$, and re-use it for all the other features. The main steps of SVTD are summed up in Algorithm \ref{alg3}.
\begin{algorithm}[h!]
\caption{SVTD}
\label{alg3}
\begin{algorithmic}[1]
\REQUIRE $M_1$, $M_2$, $M_3$, and the number of latent states $k$
\STATE Decompose ${M}_2$ as ${M}_2 = U_k S_k U_k^\top $ with a SVD.
\STATE Select a feature $r$ and compute ${M}_{3,r}$
\STATE Compute $O$ as the singular vectors of\\ $ H_r := (S_k)^{-\frac{1}{2}}U_k^\top {M}_{3,r} U_k (S_k)^{-\frac{1}{2}}$ 
\FOR{$i = 1 \to d$}
\STATE Compute $ H_i := (S_k)^{-\frac{1}{2}}U_k^\top {M}_{3,i} U_k (S_k)^{-\frac{1}{2}}$ 
\STATE Obtain the $i-$th row of $M$ as the diagonal entries of $O^\top H_i O$
\ENDFOR
\STATE Obtain $\omega$ solving $M_1 = M\omega$

\STATE Return $(M,\omega)$
\end{algorithmic}
\end{algorithm}

With this algorithm, given the values of $M_1$, $M_2$, $M_3$, one can retrieve the pair of unknown set of parameters $(M,\omega)$.
Note that step 2 of the algorithm requires to select a feature $r$ to be used to compute the matrix $O$. This matrix will be then used to diagonalize all the matrices $H_i$ in step 6. Even if theoretically one can decide to use any feature to compute $O$, it is shown by \citet{ruffini2016new} that the perturbations depend on the minimum distance between the elements of the $r-$th row of $M$, that are also the singular values of $H_r$. 
So, one repeats the step 3 of the algorithm for different features and selects the $r$ that maximizes the minimum distance between the singular values of $H_r$. An explicit implementation of a method to select the feature $r$ from which to compute the matrix $O$ is presented in Appendix \ref{sec:embed}. 
 
\section{Algorithmic details: implementing ASVTD}\label{sec:embed}

The usage of the estimates proposed in Section \ref{sec:mom} allows to modify the structure of SVTD in order to reduce the complexity and deal with perturbed matrices.  In this section, we propose a modification of the method proposed by \citet{ruffini2016new} to reach these goals. The algorithm we are going to present will produce the same results of SVTD when fed with the approximated tensors of Section \ref{sec:mom}, but in a faster way.  First, we want to exploit the nice form of the operators introduced at equations \eqref{eq:M23t} to improve the scalability of the method. Pure SVTD in fact works with all the explicit slices of the $d\times d \times d$ tensor $M_3$; we want to avoid this dependence, working with matrices with a lower dimension. Second, we need to indicate a way to select the feature $r$ from which to compute matrix $O$ of singular vectors used to perform the simultaneous diagonalization; in fact, in the unperturbed version of the theorem we can just randomly select a feature among the $d$ available. Here, with perturbed matrices, we want to do this choice in order to minimize the perturbations on the output. These modifications are presented in Algorithm \ref{alg:SVTD approx}, named Approximate-SVTD (ASVTD).

\begin{algorithm}[h!]
\caption{Approximate - SVTD}
\label{alg:SVTD approx}
\begin{algorithmic}[1]
\REQUIRE The sample $\bar{X}\in\R^{N\times d}$, the number of latent states $k$
\STATE Compute $\tilde{M}_2^{(N)}  = \frac{\bar{X}^\top \bar{X}}{N}\in \R^{d \times d}
$

\STATE Decompose $\tilde{M}_2^{(N)}$ as $\tilde{M}_2^{(N)} = U_k S_k U_k^\top $ with a SVD.
\STATE Project $\bar{X}$ on $\R^{N\times k}$: $\bar{X}_k = \bar{X}U_k(S_k)^{-\frac{1}{2}}  $
\STATE Initialize $\alpha_{min} = -\infty$ and $O = \mathbb{I}_k $ as the $k\times k$ identity matrix.
\FOR{$i = 1 \to d$}
\STATE Compute $ H_i := \frac{\bar{X}_k^\top diag((x_i^{(1)},...,x_i^{(N)})) \bar{X}_k}{N}$ 
\STATE Compute the singular values of $H_i$, $(s_1,...,s_k)$ and the left singular vectors $O_i$.
 \IF{$\min_{i\neq j}(|s_{i}-s_{j}|)>\alpha_{min}$} 
 \STATE Set: $\alpha_{min} = \min_{i\neq j}(|s_{i}-s_{j}|) $, and $O = O_i$
 \ENDIF
\ENDFOR
\FOR{$i = 1 \to d$}
\STATE Obtain the $i-$th row of $\tilde{M}$ as the diagonal entries of $O^\top H_i O$
\ENDFOR
\STATE Obtain $\tilde{\omega}$ solving $
(\frac{\sum_i x_j^{(i)}}{N})_{j=1..d} = (\tilde{M}\tilde{\omega})_{j=1..d}
$

\STATE Return $(\tilde{M},\tilde{\omega})$
\end{algorithmic}
\end{algorithm}
Comparing SVTD (Algorithm \ref{alg3}) and its approximate version (Algorithm \ref{alg:SVTD approx}), we notice that the main differences are in the loop from row 5 to 11 of Algorithm \ref{alg:SVTD approx}. In fact,
standard SVTD computes the matrices $H_i$, at its line 5, using directly the $i-th$ slice of $M_3$: 
$$
H_i = (S_k)^{-\frac{1}{2}}U_k^\top {M}_{3,i} U_k (S_k)^{-\frac{1}{2}};
$$
this operation is pretty expensive, as it requires $O(d^2k)$ operations, and is repeated $d$ times, for a total cost of $O(d^3k)$.
Instead, in ASVTD, we never compute explicitly $\tilde{M}_3^{(N)}$, but we compute directly $H_i$, exploiting the fact that, in this case
$$
H_i = \frac{((S_k)^{-\frac{1}{2}}U_k^\top \bar{X}^\top) diag((x_i^{(1)},...,x_i^{(N)})) (\bar{X} U_k (S_k)^{-\frac{1}{2}})}{N}.
$$
So, in row 3, we compute the matrix $
\bar{X}_k = \bar{X} U_k (S_k)^{-\frac{1}{2}}
$ and we use it during all the algorithm; the calculation of each $H_i$ now costs only $O (k^2N)$, which in many cases is a substantial improvement (especially when the number of the features is high). The second main difference between SVTD and ASVTD is the fact that we use the loop from row 5 to 11 of Algorithm \ref{alg:SVTD approx} to compute the value of the orthornomal matrix $O$ to be used to diagonalize the various $H_i$, selecting the feature that maximizes the difference between the two nearest singular values. It is shown in \citep[][Thm. 5.1]{ruffini2016new} that in this way, we are going to maximize the stability of the resulting matrices $(\tilde{M},\tilde{\omega})$. We now analyze the complexity requirements of ASVTD. It is easy to see that the steps from 1 to 3 have a time complexity of $O(d^2N)$ for step 1, $O(d^2k)$ for step 2, using randomized techniques \citep{RandSVD} and $ O(Ndk) $ for step 3. In the loop from step 5 to 11 we have the calculation of $H_i$, costing $O (k^2N)$, and a  $k\times k$ SVD, costing $O(k^3)$, giving a total time complexity of $ O(d^2N + dk^2N + d k^3)$ in the realistic cases  where $d,k < N$. Also the memory requirements are mild, as the only storage is for the matrices $H_i$, costing $O(dk^2)$, $\tilde{M}_2^{(N)}$, costing $O(d^2)$ and $\bar{X} $, for $ O(dN)$. The total memory is thus $ O(dk^2 + d^2 + dN)$.

\section{Experimental comparison of ASVTD with other methods}\label{sec:expcomp}

In this paper we have described how to recover the parameters $(M,\omega)$ of a NBM, presenting a simple heuristic. The advantage of our method is mainly practical, as it does not require any expert knowledge added in the preprocessing of the data.
In this paragraph we want to compare experimentally the clustering  accuracy of the proposed method with that of state-of-the-art methods. We proceed as follows: \textit{First} we generate a synthetic dataset, distributed as a mixture of independent Bernoulli, registering for each sample, the cluster to which it belongs (i.e. the latent variable generating the observation); the synthetic parameters $(M,\omega) $ have been generated as exponentially distributed random numbers, normalized to be between 0 and 1.
Then we cluster the samples in an unsupervised way, and we compare the obtained clusters with the theoretical ground-truth, registered when generating the data. We then repeat this procedure for various model and sample sizes. In order to analyze the clustering accuracy, we use the Adjusted Rand Index from \cite{hubert1985comparing} between the theoretical ground-truth partition and the one obtained with the clustering algorithm.  Adjusted Rand Index  is an indicator that compares whether two partitioning of a dataset are or not similar; an  Adjusted Rand Index of 1 indicates that two clusterings are identical, while a random labeling should have a value close to 0. Note that many well-known measures of cluster fitness such as the silhouette coefficient are not applicable as they rely on notions of distance among instances.
We run the clustering with many different algorithms:
\begin{itemize}
\item ASVTD, as described in this paper.
\item Na{\"\i}ve Bayes clustering with the three-view approach from \citet{SpectralLatent} (labeled ''HMM12'' in the figures).
\item Na{\"\i}ve Bayes clustering with the three-view approach from  \citet{TensorLatent} (labeled ''TPM'' in the figures).
\item K-Means.
\item Spectral Clustering from \citet{von2007tutorial}, using a linear kernel  ( ''SC-lin'' in the figures).

\item PCA clustering, where k-means is used to cluster the the sample where the dimensionality has been reduced usinga a PCA.
\end{itemize}

 \begin{figure*}[!ht]
              \centering
            \includegraphics[width=\textwidth]{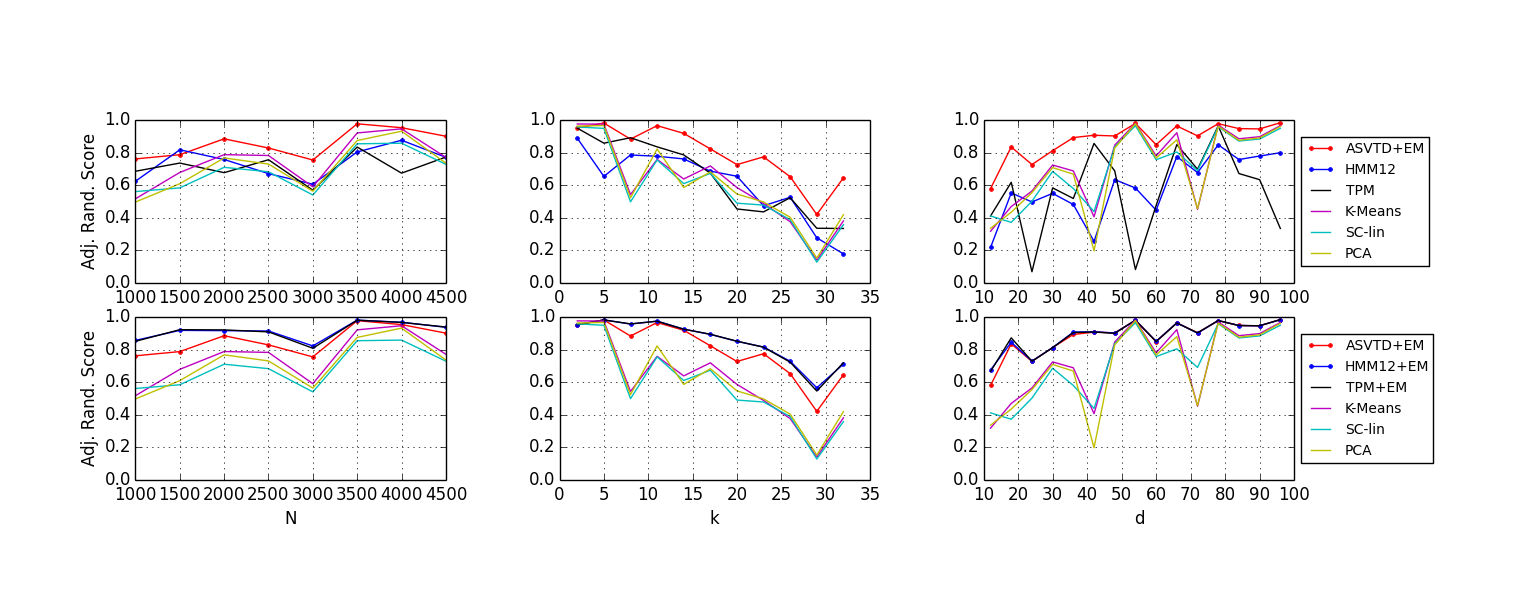}
 \caption{Adjusted Rand Index of the compared methods, when EM is not used after the multi-view methods (top three figures) and when is used (bottom three figures).}
 \label{fig:EM}
\end{figure*}

We have not run pure EM with random initialization, as its final results and running time 
strongly depend on the random initialization. For K-Means and the Spectral clustering methods we used  the implementation present in Scikit-learn Python library \citep{pedregosa2011scikit}. In order to avoid ordering issues for the multi-view clustering methods, we just used the true parameters of the models to recover the correct ordering; the views have been built by dividing the set of the features into three subsets: the first view are the features from $1$ to $d/3$, the second those from $d/3+1$ to $d2/3$ and the last are those from $d2/3+1$ to $d$; this was an easy task, as the entries of $M$ are well balanced. After ASVTD, we always run EM until convergence is reached, stopping the algorithm when the norm of the variation on the estimated $\omega$ between an iteration and the next was less than $0.01$; we always use EM, because ASVTD is an approximate method, without theoretical guaranteers, and EM is an integral part of its usage. In a first experiment we do not run any EM iteration after the multi-view clustering methods, while in the second one, we do run EM after those methods, with the same stopping criteria used for ASVTD\footnote{All the algorithms have been implemented in Python 2.7, using \textit{numpy} \citep{van2011numpy} library for linear algebra operations. All the experiments have run on a MacBook Pro, with an Intel Core i5 processor. }. 

The results are displayed in figure \ref{fig:EM}. The top three figures represent the experiment where EM is not run after the  multi-view clusterings, while the three figures below represent the experiment where EM is used. In the leftmost figures we fixed the number of features $d=99$ and the number of latent states $k=12$, varying the sample size from $N=100$ to $N = 10000$. In the central figures, the sample size $N=10000$ and the number of features $d=99$ are fixed, while the number of latent states varies from $k=2$ to $k=33$. In the rightmost figure, the sample size $N=10000$ and the number of latent states $k=4$ are fixed, while the number of features varies from $d=12$ to $d=99$. It is interesting to notice that all the reference methods (K-means, Spectral Clustering and PCA) are outperformed from tensor methods; also, seems that the intuition of using a mixed tensor + EM method provides good results, as ASVTD, despite the lack of theoretical guarantees, performs better than the pure (i.e. without EM) multi-view approaches, both with Tensor Power method and with  the method from \citep{SpectralLatent}. If we plug some EM iterations after the pure tensor methods (three figures below), we can see interesting results. EM provides a huge improvement to the quality of the results of TPM and of \citet{SpectralLatent} method. In particular, we can see that now ASVTD gets almost the same results as these other methods, without requiring any domain-dependent preprocessing of the data; in this case, this preprocessing activity was trivial, but, as explained in paragraph \ref{sec:other methods}, with certain unbalanced datasets, it may be very challenging to find a proper splitting of the features. In addition, if we look at the experiments of Section \ref{sec:patient clustering}, we can see that they are in the case where $N$ is large, $d$ is large as well and $k$ is small; these are exactly the configurations where ASVTD performs better, providing results at least as good as the competing methods; these motivations justify our choice of using ASVTD for the experimental results of Section  \ref{sec:patient clustering}.

 \begin{figure}[h]
              \centering
            \includegraphics[width=9cm]{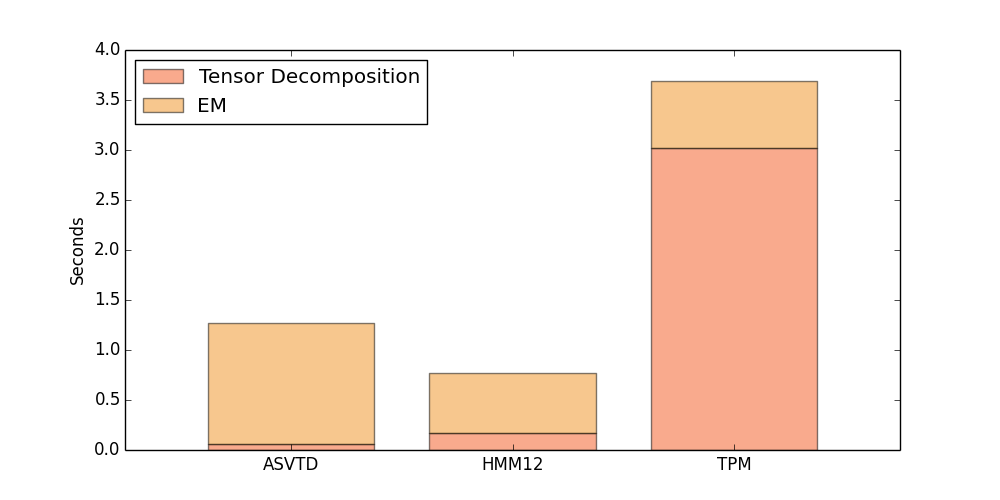}
 \caption{Average running times of the tensor-based methods. The red portion of the bars represents the average time spent in doing tensor decomposition; the orange portion, the time for EM.}
 \label{fig:Time}
\end{figure}
We now briefly discuss the running times. For each run of the experiment described in the previous paragraph we registered the time employed by each of the clustering methods. We then average all these records, in order to get, for each method, the average time needed to perform the clustering. Results are displayed in Table \ref{tab:time}.
\begin{table}[!t]
\centering
\begin{tabular}{c c}
\hline
\textbf{Method} & \textbf{Time (seconds)}  \\\hline
ASVTD + EM &1.3\\\hline
TPM + EM & 3.68 \\\hline
HMM12 + EM& 0.77\\\hline
KMeans & 0.76\\\hline
SC - lin &  7.48\\\hline
PCA & 0.38\\\hline
\end{tabular}
\caption{The average clustering time spent for each method.}
\label{tab:time}
\end{table}
It is immediate to see that Spectral Clustering method is much slower than the competing techniques. This is a consequence of dealing with an $N\times N$ affinity matrix, that requires a number of operations quadratic in the sample size. PCA and K-means are pretty fast, but we have seen that they perform poorly on binary data.
Looking at tensor methods, TPM seems to be the slowest method, consequence of a worst scalability on the number of latent components, while ASVTD and HMM12 have similar performances. Focusing on the tensor based methods, it is interesting to understand the amount of time spent in performing tensor decomposition and the amount spent in refining parameters with EM, to reach convergence. This analysis is displayed in Figure \ref{fig:Time}; the red portion of each bar represents the time for decomposition, the orange portion, the time for EM; the total bars are the average time spent by each algorithm to perform the full reconstruction (decomposition and EM).
As expected, ASVTD is significantly faster than other decomposition methods, due to the better scalability. However, as it does not provide a guaranteed estimation of the model parameters, EM takes a bit more time to converge. If we look at the other algorithms we can see a significantly longer time in decomposing tensors, and a shorter time in doing EM. 

\end{document}